\documentclass[a4paper]{article}
\usepackage{INTERSPEECH2016,amssymb,amsmath,epsfig}
\ninept

\def\reg{{\rm\ooalign{\hfil
     \raise.07ex\hbox{\scriptsize R}\hfil\crcr\mathhexbox20D}}}
\title{Small-footprint Deep Neural Networks with Highway Connections \\ for Speech Recognition}

\makeatletter
\def\name#1{\gdef\@name{#1\\}}
\makeatother \name{{\em Liang Lu, Steve Renals}
\thanks{Funded by the EPSRC Programme Grant EP/I031022/1, Natural Speech Technology (NST). The NST research data collection may be accessed at http://datashare.is.ed.ac.uk/handle/10283/786. We thank Yu Zhang and Dong Yu for helpful discussions on using the CNTK toolkit.}}

\address{Centre for Speech Technology Research, The University of Edinburgh, Edinburgh, UK \\
{\small \tt \{liang.lu, s.renals\}@ed.ac.uk }}

\begin{document}
%\ninept
%
\maketitle
\begin{abstract}
For speech recognition, deep neural networks (DNNs) have significantly improved the recognition accuracy in most of benchmark datasets and application domains. However, compared to the conventional Gaussian mixture models, DNN-based acoustic models usually have much larger number of model parameters, making it challenging for their applications in resource constrained platforms, e.g., mobile devices. In this paper, we study the application of the recently proposed highway network to train small-footprint DNNs, which are {\it thinner} and {\it deeper}, and have significantly smaller number of model parameters compared to conventional DNNs. We investigated this approach on the AMI meeting speech transcription corpus which has around 80 hours of audio data. The highway neural networks constantly outperformed their plain DNN counterparts, and the number of model parameters can be reduced significantly without sacrificing the recognition accuracy. 

\end{abstract}
\noindent{\bf Index Terms}: speech recognition, highway network, small-footprint deep learning.

\section{Introduction}
\label{sec:intro}

Modern state-of-the-art speech recognition systems are based on neural network acoustic models~\cite{hinton2012deep, dahl2012context, bourlard1994connectionist, morgan1995neural, renals1994connectionist}. A typical architecture is the deep neural network (DNN)~\cite{hinton2012deep, dahl2012context}, which is a feedforward neural network with multiple hidden layers (e.g., 4 $\sim$ 9), and each layer has a large number of hidden units (e.g., 512 $\sim$ 2048). Compared to the conventional Gaussian mixture models, DNN acoustic models usually have much larger number of model parameters, which explains their large statistical modelling capacities and high recognition accuracies. However, it becomes challenging for the applications of DNN-based speech recognition systems in resource constrained scenarios. For instance, it is highly desirable that the speech recognition system can still function in wearable computing and mobile devices when the internet connection is unavailable. This requires that smaller size of acoustic models can still achieve high recognition accuracy. 

There have been a number of works on small footprint DNNs for this purpose. For instance, Xue et al.~\cite{xue2013restructuring} and Sainath et al.~\cite{sainath2013low} approximate the weight matrix between two hidden layers by a product of two low-rank matrices, which may be equivalent to insert a bottleneck layer in between without the nonlinear activation. Another branch of studies are based on the {\it teacher-student} architecture~\cite{li2014learning, ba2014deep, romero15_fitnet}, which is also referred to as model compression~\cite{bucilu2006model} and knowledge distillation~\cite{hinton2015distilling}. In this approach, the {\it teacher} may be a large-size network or an ensemble of several different models, which is used to predict the soft targets for training the {\it student} model that is much smaller. As discussed in~\cite{hinton2015distilling}, the soft targets provided by the teacher encode the generalisation power of the teacher, and the student model trained using these labels is observed to perform better than a small model trained in the usual way~\cite{li2014learning, ba2014deep, romero15_fitnet}. Recently, \cite{sindhwani2015structured} investigated the use of low rank displacement of structured matrices (e.g., Teoplitz matrix) for small-footprint neural networks. This work is in line with the argument that neural networks with dense connections are over-parameterised, and the linear layer may be replaced by structured efficient linear layers (SELLs)~\cite{le2013fastfood, yang2014deep, moczulski2015acdc}.

In this paper, we investigate the {\it thin} and {\it deep} architectures for small-footprint neural network acoustic models. However, as the depth increases, training DNNs by stochastic gradient decent (SGD) becomes increasingly difficult due to the highly non-convexity of the error surface. One approach is to pre-train the neural network by unsupervised ~\cite{hinton2006reducing} or greedy layer-wise fashion~\cite{bengio2007greedy}. However, this approach cannot circumvent the difficulty arises in the fine tuning stage. Another approach is to rely on the {\it teacher-student} architecture, e.g. the FitNet~\cite{romero15_fitnet}, but it requires the additional effort to train the teacher model beforehand. Our work in this paper builds on the recently proposed highway networks~\cite{srivastava2015training}, where the {\it transform} gate is used to scale the output of a hidden layer and the {\it carry} gate is used to pass through the input directly after elementwise rescaling. Similar idea has also been studied on long short-term memory recurrent neural networks (LSTM-RNN) for speech recognition~\cite{zhang2015highway}. In this work, we observe that the highway connections can be successfully applied to training {\it thinner} and {\it deeper} networks, while still retraining the recognition accuracy. Our experiments were performed on the AMI meeting speech transcription corpus, which contains around 70 hours of training data. Using highway neural networks, we managed to cut down the number of model parameters by over 80\% with marginal accuracy loss compared to our baseline DNN acoustic models. %In addition, we also evaluated the role of the transform and carry gate respectively. 

\section{Highway Deep Neural Network}

\subsection{Deep neural networks}
A DNN is a feed-forward neural network with multiple hidden layers that performs cascaded layer-wise nonlinear transformations of the input. For a network with $L$ hidden layers, the model may be represented as
\begin{align}
\mathbf{h}_1 &= f(\mathbf{x}, \theta_1) \\
\mathbf{h}_l &= f(\mathbf{h}_{l-1}, \theta_l), \quad \text{for} \quad l=2,\ldots, L \\
\mathbf{y} &= g(\mathbf{h}_L, \varphi)
\end{align}
where $\mathbf{x}$ is an input vector to the network; $f(\mathbf{h}_{l-1}, \theta_l)$ denotes the transformation of the input $\mathbf{h}_{l-1}$ with the parameter $\theta_l$ followed by a nonlinear activation function (e.g., {\tt sigmoid} or {\tt tanh}); $g(\cdot, \varphi)$ is the output function(e.g. {\tt softmax}) which is parameterised by $\varphi$ in the output layer. Given the ground truth target $\hat{\mathbf{y}}$, the network is usually trained by gradient decent to minimise a loss function $\mathcal{L}(\mathbf{y}, \hat{\mathbf{y}})$ (e.g. cross-entropy). However, as the number of hidden layers increases, the error surface becomes increasingly non-convex, and it is more possible to find a poor local minima using gradient-based optimisation algorithms with random initialisation~\cite{erhan2009difficulty}. Furthermore, \cite{glorot2010understanding} showed that the variance of the back-propagated gradients may become small in the lower layers if the model parameters are not initialised properly. 

%Pre-training (unsupervised or supervised layer-wise) has been shown to important for training deep networks, however, pretrained deeper networks empirical results (including this work) suggest that with more training data, and for deeper network (e.g. $L >10$), deeper networks initialised by pre-training still perform worse. 

\subsection{Highway networks}

There have been numerous studies on overcoming the difficulties in training very deep neural networks, including pre-training~\cite{hinton2006reducing, bengio2007greedy}, normalised initialisation~\cite{glorot2010understanding}, deeply-supervised networks~\cite{lee2014deeply}, etc. Recently, Srivastav et al.~\cite{srivastava2015training} proposed the highway network and demonstrated good results to train very deep networks (e.g., up to 100 hidden layers). In the highway network, the hidden layers are augmented with two gating functions, which can be represented as
\begin{align}
\label{eq:hw}
\mathbf{h}_l = f(\mathbf{h}_{l-1}, \theta_l)\circ T(\mathbf{h}_{l-1}, \mathbf{W}_T) \nonumber \\
+ \mathbf{h}_{l-1}\circ C(\mathbf{h}_{l-1}, \mathbf{W}_c)
\end{align}
where $T(\cdot)$ is the {\it transform} gate that scales the original hidden activations; $C(\cdot)$ is the {\it carry} gate, which scales the input before passing it directly to the next hidden layer; $\circ$ denotes elementwise (Hadamad) product; The outputs of $T(\cdot)$ and $C(\cdot)$ are constrained to be $[0, 1]$, and we use sigmoid functions for both gates parameterised by $\mathbf{W}_T$ and $\mathbf{W}_c$ respectively. Unlike~\cite{srivastava2015training}, in this work, we do not use any bias vector in the two gate functions. In~\cite{srivastava2015training}, the carry gate is constrained to be $C(\cdot) = \mathbf{1} - T(\cdot)$, while in this work, we evaluate the generalisation ability of highway networks with and without this constraint. 

Without the transform gate, i.e. $T(\cdot) = \mathbf{1}$, the highway network is similar to a network with skip connections -- the main difference is that the input is firstly scaled by the carry gate. Without the carry gate,  i.e. $C(\cdot) = \mathbf{0}$, the hidden layer is
\begin{align}
\mathbf{h}_l = f(\mathbf{h}_{l-1}, \theta_l)\circ T(\mathbf{h}_{l-1}, \mathbf{W}_T).
\end{align}
At first glance, it looks similar to the dropout regularisation for neural networks~\cite{hinton2012improving}, which may be represented as
\begin{align}
\mathbf{h}_l = f(\mathbf{h}_{l-1}, \theta_l)\circ \pmb{\epsilon}, \quad \epsilon_i \sim p(\epsilon_i),
\end{align}
where $p(\epsilon_i)$ is a Bernoulli distribution for each element in $\pmb{\epsilon}$ as originally proposed in~\cite{hinton2012improving}, while it was shown later that using a continuous distribution with well designed mean and variance works as well or better~\cite{srivastava2014dropout}. From this perspective, the transform gate may work as a regulariser, but with the key difference that $T(\cdot)$ is a deterministic function, while $\epsilon_i$ is drawn stochastically from a predefined distribution in dropout. Nevertheless, our empirical results (cf. Section~\ref{sec:gate}) indicate that the transform gate and the carry gate can speed up the convergence rate. In addition, the highway networks also generalise better when measured in terms of recognition accuracy, which is presumably due to the regularisation effect of the two gating functions.

\subsection{Small-footprint networks}

The aim of this paper is to train small-footprint neural networks for resource constrained speech recognition. From Eq.~\eqref{eq:hw}, the highway network is not directly suitable for this purpose, because it introduces additional computational cost and model parameters in the two gating functions. The rationale is that the computational complexity and the number of model parameters for each layer in a densely connected network are in the order of $O(n^2)$, where $n$ is the size of hidden units. Increasing the depth of the network only linearly increases the computational cost and the model size, while reducing the width can yield the quadratic reduction in the two metrics. Highway connections make it feasible to train very {\it thin} and {\it deep} networks, and therefore the overall model sizes are much smaller. To further save the model parameters, in this work, we shared the two gates for all hidden layers so that the additional number of model parameters for $T(\cdot)$ and $C(\cdot)$ is relatively small.

\begin{figure}[t]
\small
\centerline{\includegraphics[width=0.25\textwidth]{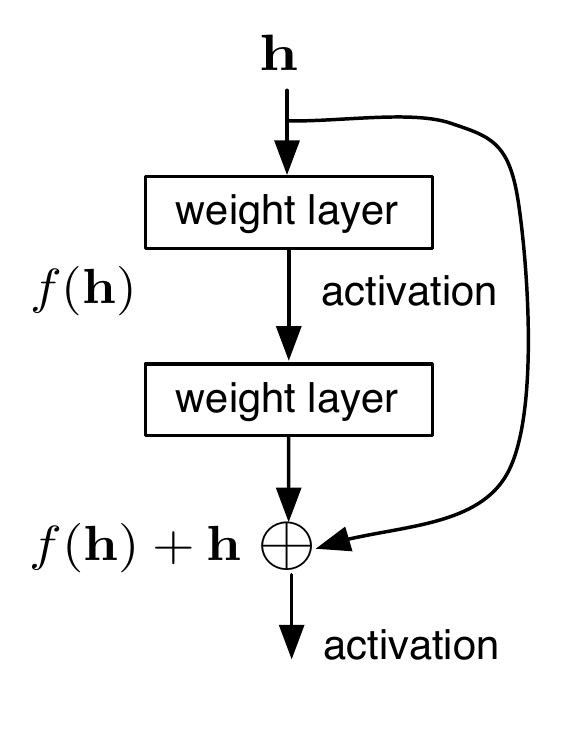}} \vskip -4mm
\caption{The building block of residual networks~\cite{he2015deep} }  
\label{fig:renet}
\vskip -4mm
\end{figure}

\subsection{Comparison to residual networks}

Residual network is a type of very deep network using skip connections, which has achieved state-of-the-art results in image recognition~\cite{he2015deep}.  The building block for residual networks is shown in Figure \ref{fig:renet}. In fact, residual networks are similar to highway networks without the two additional gate functions, which can significantly reduce the computational cost. It also reduces the number of model parameters, albeit the reduction is marginal because the two gating functions are tied for all the hidden layers in our configuration. However, without the gating functions, training residual networks may be more difficult compared to highway networks, which will be empirically studied in the following experimental section.

 \begin{table}
 \centering \small
\caption{Comparison of depth and width between plain DNNs and HDNNs. $^*$indicates that the models were trained using the Kaldi toolkit, where the networks were initialised with restricted Boltzmann machine (RBM) based pre-training because random initialisation did not yield convergence.}\vskip 1.5mm

\label{tab:depth}
\begin{tabular}{l|cccc}
\hline 

\hline
System & \#Layer & Dim & \#Parm (M) & WER  \\ \hline
{\footnotesize GMM+SAT+bMMI} & - & - &6.48 & 31.7 \\
%DNN & 4 & 2048 & 21.9 & \\
DNN & 6   & 2048 & 30.3 & 26.8 \\
DNN & 6   & 1024 & 9.9 &  27.2 \\
%DNN & 6   & 1024 & 9.9 &  27.2 \\
DNN & 10 & 2048 & 47.1 & 27.7 \\
DNN & 10 & 1024 & 14.1 & 27.9 \\
DNN$^*$ & 10 & 512 & 4.7 & 28.8 \\
DNN$^*$ & 10 & 256 & 1.8 & 31.5 \\
DNN$^*$ & 15 & 1024  & 19.4 & 27.6\\
DNN$^*$ & 15 & 512 & 6.0 & 29.1 \\ 
DNN$^*$ & 15 & 256 & 2.1 & 31.5 \\ \hline
HDNN & 10 & 2048 & 55.5& 26.8 \\ 
HDNN & 10 & 1024 & {\bf 16.2} & {\bf 26.8} \\
HDNN & 10 & 512 & 5.2 & 27.2 \\
HDNN & 10 & 256 & 1.9 & 28.8 \\
HDNN & 10 & 128 & 0.77 & 32.0 \\ \hline
HDNN & 15 & 1024 & 21.5 & 26.8 \\
HDNN & 15 & 512 & {\bf 6.5} & {\bf 27.1} \\
HDNN & 15 & 256 & 2.2 & 28.5 \\
HDNN & 15 & 128 & {\bf 0.85} & {\bf 31.4} \\ 
\hline

\hline
\end{tabular}
\vskip-4mm
\end{table}

\section{Experiments}
\label{sec:exp}

\subsection{System setup}

Our experiments were performed on the individual headset microphone (IHM) subset of the AMI meeting speech transcription corpus~\cite{renals2007recognition}. The amount of training data is around 80 hours, corresponding to roughly 28 million frames. This dataset is much larger than most of the datasets (e.g. MNIST, CIFAR, etc.) where other types of thin and deep networks were evaluated~\cite{romero15_fitnet, srivastava2015training}. We used 40-dimensional fMLLR adapted features vectors normalised on per-speaker level, which were then spliced by a context window of 15 frames (i.e. $\pm7$) for all the systems. The number of tied HMM states is 3972, and all the DNN systems were trained with the same alignment. The results reported in this paper were obtained using the CNTK toolkit~\cite{yu2014introduction} with the Kaldi decoder~\cite{povey2011kaldi}, and the networks were trained using the cross-entropy (CE) criterion without pre-training unless specified otherwise. We set the momentum to be 0.9 after the 1st epoch, and we used the sigmoid activation for the hidden layers. The weights in each hidden layer were randomly initialised with a uniform distribution in the range of $[-0.5, 0.5]$ and the bias parameters were initialised to be $0$ for CNTK systems. We used a trigram language model for decoding. 

\begin{figure}[t]
\small
\centerline{\includegraphics[width=0.5\textwidth]{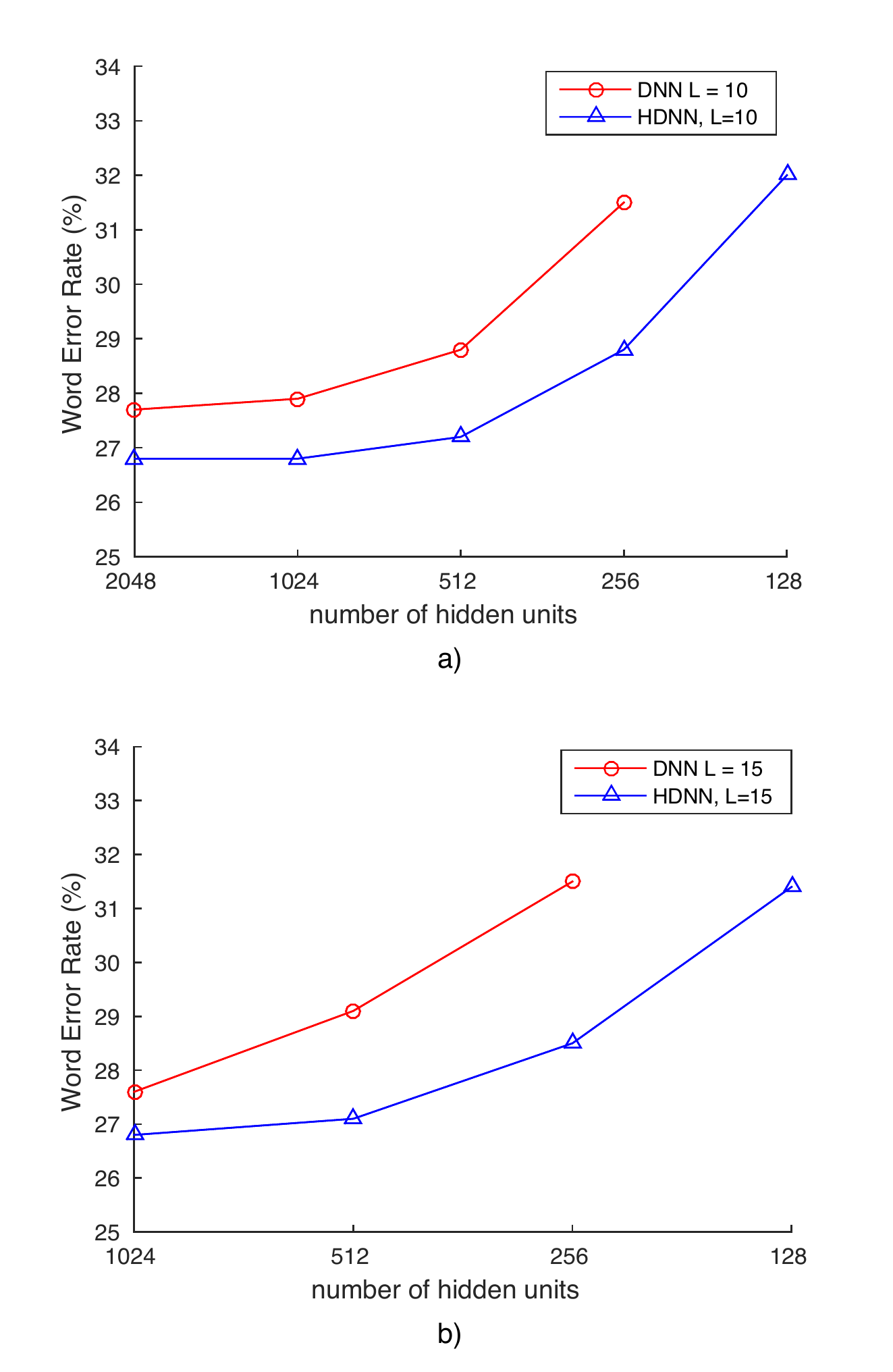}} \vskip -4mm
\caption{ Comparison between plain DNNs and HDNNs with different number of hidden units. Thin and deep HDNNs achieved consistent lower WERs than their plain DNN counterparts. }  
\label{fig:depth}
\vskip -2mm
\end{figure}

\subsection{Depth vs. Width}

Table \ref{tab:depth} shows the word error rates (WERs) of plain DNNs and highway networks (HDNNs) with different configurations. As the number of hidden units decreases, the accuracy of plain DNNs degrade rapidly, and the accuracy loss cannot be compensated by increasing the depth of the network. We faced the difficulty to train thin and deep networks directly without RBM pre-training (the CE loss did not decrease at all after many epochs). However, with highway connections we did not have this difficulty. The HDNNs achieved consistent lower WERs compared to the plain DNN counterparts, and the margin of the gain also increases as the number of hidden units becomes smaller as shown in Figure \ref{fig:depth}. With highway connections, we can cut down the number of model parameters by around 80\% with marginal accuracy loss, and with less than 1 million model parameters, the CE trained HDNN can achieve comparable or slight higher accuracy compared to a strong GMM baseline with speaker adaptive training (SAT) and bMMI-based discriminative training. The accuracy of smaller-size HDNN models may be further improved by the teacher-student style training, which will be investigated in the future. 

\subsection{Transform gate vs. Carry gate}
\label{sec:gate}

\begin{figure}[t]
\small
\centerline{\includegraphics[width=0.5\textwidth]{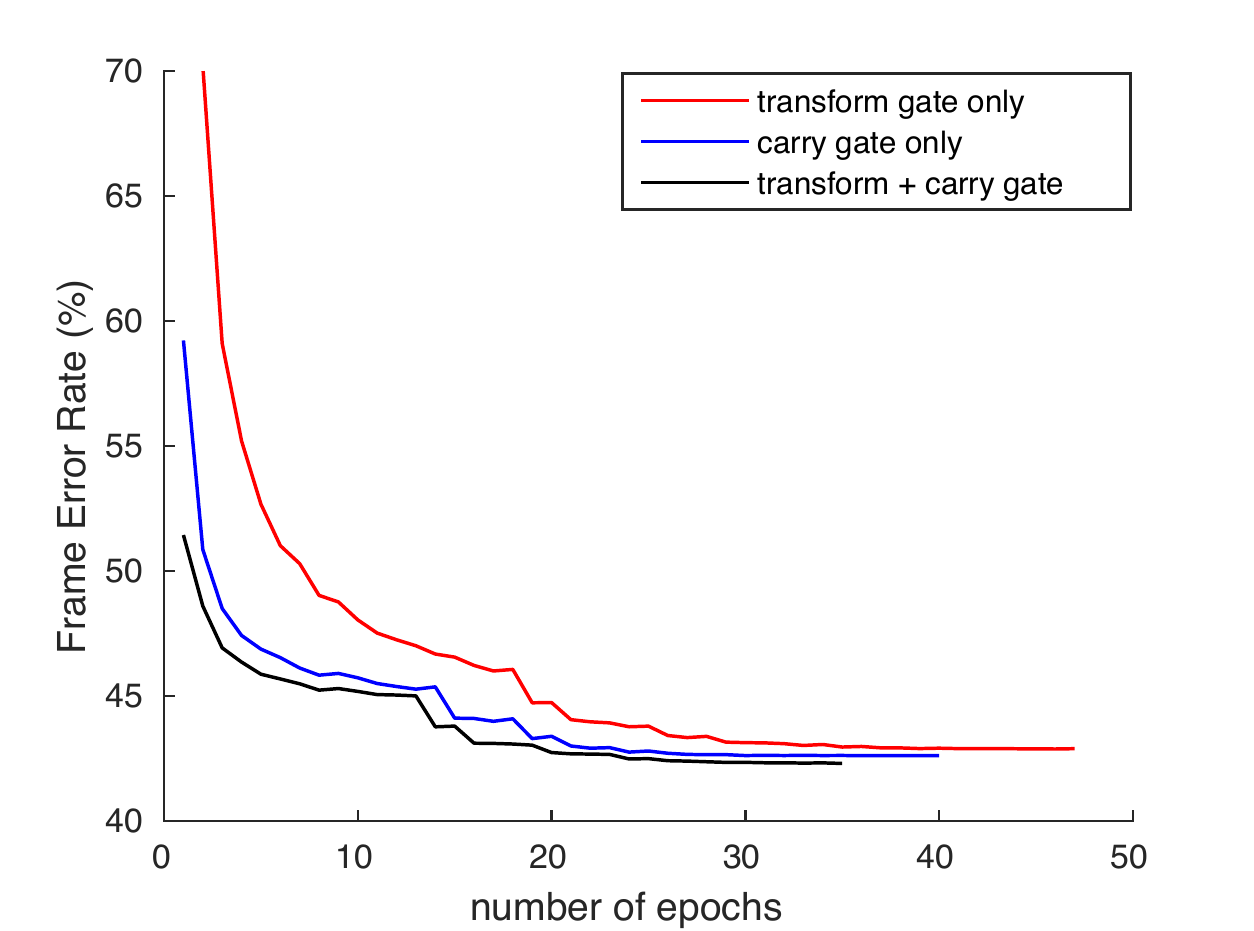}} \vskip -2mm
\caption{The convergence curve of training HDNNs with and without the transform and carry gate. The Frame Error Rates (FERs) were obtained from the validation dataset.}  
\label{fig:gate}
\end{figure}

We then evaluated the specific role of the transform and carry gate in the highway architectures. The results are shown in Table \ref{tab:gate}, where we disabled one of both of the gates. We observed that using only one of the two gates, the HDNN can still achieved lower WER compared to the plain DNN, but the best results were obtained when both of the gates were active, which indicates that the two gating functions are complementary to each other. Figure \ref{fig:gate} shows the convergence curve of training HDNNs with and without the transform and carry gate. We observed that it converged faster when both of the gates were turned on. With only the transform gate, the convergence rate was much slower.  As discussed before, the carry gate can be viewed as a particular type of skip connection, and it was more important to speed up the convergence compared to the transform gate in our experiments. 

\subsection{Constrained carry gate}

We also evaluated using the constrained carry gate in our experiments, where $C(\cdot) = \mathbf{1} - T(\cdot)$ as studied in~\cite{srivastava2015training}. In this approach, the computational cost is reduced since the matrix-vector multiplication for the carry gate is not required. We evaluated this configuration with 10-layer neural networks, and the results are shown in Table \ref{tab:carry}. Contrary to our expectations, with the constrained carry gate only we obtained worse results when the networks were relatively wide, while the accuracy gap was reduced when the number of hidden units was smaller. The reason may be that in the constrained setting, the transform gate $T(\cdot)$ learns the scaling function for both the input and output at the same time. As regularisation is expected to be more important for training wide and deep networks, this may not be achieved by using a single gating function. For instance, both the input and output of one hidden layer may require larger or smaller scaling weights at the same time, which is impossible in the constrained setting. In the future, we shall look into the regularisation and generalisation properties of the two gating functions more closely.    

 \begin{table}
\caption{Results of highway networks with and without the transform and carry gate. }\vskip 1mm
\label{tab:gate}
\centering \small
\begin{tabular}{l|ccccc}
\hline 

\hline
System & \#Layer & Dim & Transform & Carry & WER  \\ \hline
DNN$^*$    & 10 & 512 & $\times$ & $\times$ & 28.8 \\
HDNN & 10 & 512 & $\surd$ & $\surd$ & 27.2 \\
HDNN & 10 & 512 & $\surd$ & $\times$ & 27.6 \\
HDNN & 10 & 512 & $\times$ & $\surd$ & 27.5 \\
\hline

\hline
\end{tabular}
\vskip-3mm
\end{table}

 \begin{table}
\caption{Results of using constrained carry gate, where $C(\cdot) = \mathbf{1} - T(\cdot)$. }\vskip 1mm
\label{tab:carry}
\centering \small
\begin{tabular}{l|cccc}
\hline 

\hline
System & \#Layer & Dim & Constrained & WER  \\ \hline
DNN    & 10 & 1024 & - & 27.9 \\
HDNN & 10 & 1024 & $\times$ & 26.8 \\
HDNN & 10 & 1024 &  $\surd$ & 28.0 \\ \hline
DNN$^*$ & 10 & 512 & - & 28.8 \\
HDNN & 10 & 512 &  $\times$ & 27.2 \\
HDNN & 10 & 512 &  $\surd$ & 27.4 \\ \hline
DNN$^*$ & 10 & 256 & - & 31.5 \\
HDNN & 10 & 256 &  $\times$ & 28.8 \\
HDNN & 10 & 256 &  $\surd$ & 29.6 \\ 
\hline

\hline
\end{tabular}
\vskip-3mm
\end{table}

 \begin{table}[t]
 \centering \small
\caption{Comparison to residual networks (ResNets).}\vskip 1.5mm

\label{tab:compare}
\begin{tabular}{l|cccc}
\hline

\hline
System & \#Layer & Dim & Activation & WER  \\ \hline
DNN$^*$ & 10 & 1024 & Sigmoid & 27.9 \\ 
DNN$^*$ & 10 & 512 & Sigmoid & 28.8 \\
DNN$^*$ & 10 & 256 & Sigmoid & 31.5 \\ \hline
ResNet & 10 & 1024 & Sigmoid & 27.6 \\ 
ResNet & 10 & 512 & Sigmoid & 27.8 \\
ResNet & 10 & 256 & Sigmoid & 29.5 \\ \hline
HDNN & 10 & 1024 & Sigmoid &  26.8 \\
HDNN & 10 & 512 & Sigmoid &  27.2 \\
HDNN & 10 & 256 & Sigmoid & 28.8 \\ \hline
ResNet & 10 & 1024 & ReLU & 27.2 \\ 
ResNet  & 10 & 512 & ReLU & 27.3 \\
ResNet  & 10 & 256 & ReLU & 28.6\\ \hline
ResNet & 15 & 1024 & ReLU & 26.9 \\
ResNet & 15 & 512 & ReLU & 27.0 \\
ResNet & 15 & 256 & ReLU & 28.2 \\ \hline
HDNN & 15 & 1024 & ReLU &  27.1  \\
HDNN & 15 & 512 & ReLU &  27.3  \\
HDNN & 15 & 256 & ReLU & 28.7 \\ 
\hline

\hline
\end{tabular}
\end{table}

\subsection{Comparison to residual networks}

Finally, we compare highway networks to residual networks, and the results are given in Table \ref{tab:compare}. Our experiments showed that without the two gating functions, training the residual networks was comparably more challenging. For instance, with 10 hidden layers and using sigmoid activations, residual networks achieved higher WER compared to highway networks. However, the differences in terms of the accuracy were smaller when using ReLU (rectified linear unit) activations for residual networks, because training ReLU networks are relatively less difficult. Furthermore, we experienced difficulty to train residual networks with 15 hidden layers using sigmoid activations instead of ReLU (The CE cost did not come down after over 20 epochs), although with ReLU activations, residual networks slightly outperformed highway networks in this case. Note that, residual networks still performed better compared to the plain networks with RBM pre-training, e.g., when the depth was 10. From our experiments, we may draw the conclusion that residual networks are more powerful to train deeper networks compared to plain DNNs, particular with ReLU activation functions which reduce the optimisation difficulty. However, highway networks are more flexible with the activation functions due to the two gating functions that control the follow of information.

\section{Conclusions}

In this paper, we investigate thin and deep neural networks for small-footprint acoustic models. Our study is build on the recently proposed highway neural network, which introduces an additional transform and carry gate for each hidden layer. Our experiments indicate that the highway connections can facilitate the information flow and mitigate the difficulty in training very deep feedforward networks. The thin and deep architecture with highway connections achieved consistently lower WERs compared to plain DNNs, and by reducing the number of hidden units, we can significantly cut down the total number of model parameters with negligible accuracy loss. We also evaluated the specific role of the transform and carry gate, and we found that the carry gate was more important to speed up the convergence in our experiment. The small-footprint highway networks may be further improved by the teacher-student style training, which will be investigated in our future work.

%\newpage
\ninept
\bibliographystyle{IEEEtran}
\bibliography{bibtex}

\end{document}